\documentclass{article}
\usepackage[nonatbib,preprint]{neurips_2021}
\usepackage{pdfpages}

\title{Numerical influence of ReLU'(0) on backpropagation}
\author{%
  David Bertoin \\
  IRT Saint Exupéry\\
  ISAE-SUPAERO\\
  ANITI\\
  Toulouse, France \\
  \texttt{david.bertoin@irt-saintexupery.com}\\
  \And
  Jérôme Bolte \\
  Toulouse School of Economics\\
  Université Toulouse 1 Capitole\\
  ANITI\\
  Toulouse, France \\
  \texttt{jbolte@ut-capitole.fr}\\
  \And
  Sébastien Gerchinovitz \\
  IRT Saint Exupéry\\
  Institut de Mathématiques de Toulouse\\
  ANITI\\
  Toulouse, France \\
  \texttt{sebastien.gerchinovitz@irt-saintexupery.com}\\
  \And
  Edouard Pauwels \\
  CNRS\\
  IRIT, Université Paul Sabatier\\
  ANITI\\
  Toulouse, France \\
  \texttt{edouard.pauwels@irit.fr}\\
  }

\usepackage{amsmath,amssymb,amsthm,amsfonts}
\usepackage{float} 
\usepackage{hyperref}       
\usepackage{graphicx}
\usepackage{caption}
\usepackage{subcaption}
\usepackage{xcolor}
\usepackage{url,soul}

\newcommand{\PP}{\mathbb{P}}
\newcommand{\RR}{\mathbb{R}}
\newcommand{\NN}{\mathbb{N}}
\newcommand{\backprop}{\mathrm{backprop}}

\newcommand{\ReLU}{\mathrm{ReLU}}
\newcommand{\concatenate}{\mathrm{concatenate}}
\newcommand{\softmax}{\mathrm{softmax}}
\newcommand{\logexp}{log-exp}

\newtheorem{theorem}{Theorem}

\newtheorem{definition}{Definition}

\newcommand{\ed}[1]{\textcolor{blue}{#1}}

\begin{document}
\includepdf[pages=-]{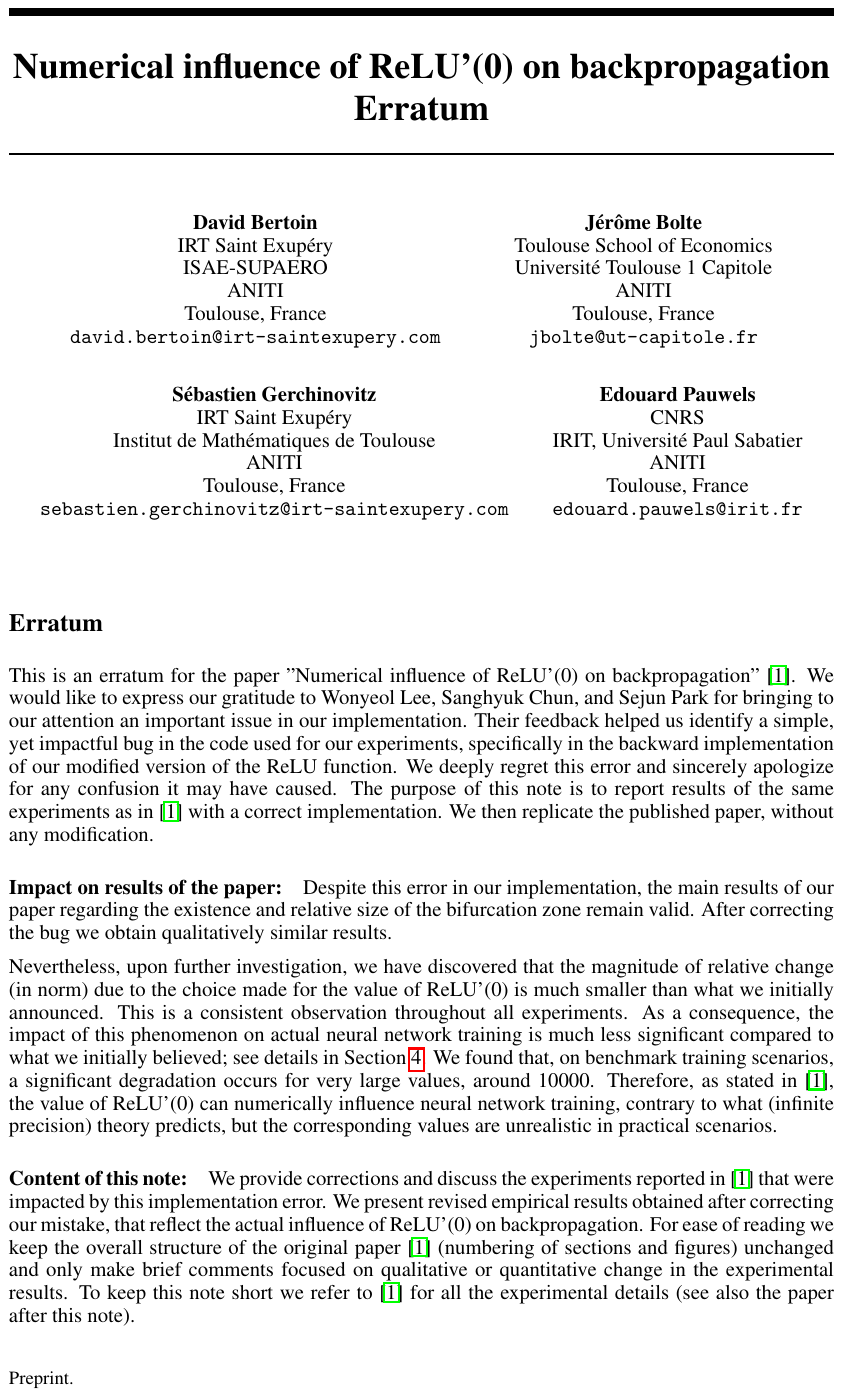}
\maketitle

\begin{abstract} 
  In theory, the choice of $\ReLU'(0)$ in $[0,1]$ for a  neural network has a negligible influence both on backpropagation and training. Yet, in the real world, 32 bits default precision combined with the size of deep learning problems makes it a hyperparameter of training methods. 
  We investigate the importance of the value of $\ReLU'(0)$ for several precision levels (16, 32, 64 bits), on various networks (fully connected, VGG, ResNet) and datasets (MNIST, CIFAR10, SVHN, ImageNet). We observe considerable variations of backpropagation outputs which occur around half of the time in 32 bits precision. The effect disappears with double precision, while it is systematic at 16 bits. For vanilla SGD training, the choice $\ReLU'(0)=0$ seems to be the most efficient. For our experiments on ImageNet the gain in test accuracy over $\ReLU'(0)=1$ was more than 10 points (two runs). We also evidence that reconditioning approaches as batch-norm or ADAM tend to buffer the influence of  $\ReLU'(0)$'s value. Overall, the message we  convey is that  algorithmic differentiation of nonsmooth problems potentially hides parameters that could be tuned advantageously.
\end{abstract}

\section{Introduction}

\paragraph{Nonsmooth algorithmic differentiation:}


The training phase of neural networks relies on first-order methods such as Stochastic Gradient Descent (SGD) \cite{goodfellow2016deep,bottou2018optimization} and crucially on algorithmic differentiation \cite{griewank2008evaluating}. The fast ``differentiator" used in practice to compute mini-batch descent directions is the backpropagation algorithm \cite{rumelhart1986learning,baydin2018automatic}. Although designed initially for differentiable problems, it is  applied indifferently to smooth or nonsmooth networks. In the nonsmooth case this requires surrogate derivatives at the non regularity points. We focus on the famous $\ReLU:=\max(0,\cdot)$, for which a value for $s = \ReLU'(0)$ has to be chosen. A priori, any value in $[0,1]$ bears a variational sense as it corresponds to a subgradient \cite{rockafellar2009variational}. Yet in most libraries $s = 0$ is chosen; it is the case for TensorFlow \cite{abadi2016tensorflow}, PyTorch \cite{NEURIPS2019_9015} or Jax \cite{jax2018github}. Why this choice? What would be the impact of a different value of $s$? How this interacts with other training strategies? We will use the notation $\backprop_s$ to denote backpropagation implemented with $\ReLU'(0) = s$ for any given real number $s$.\footnote{Definition in Section~\ref{sec:mathAutoDiff}. This can be coded explicitly or cheaply emulated by $\backprop_0$. Indeed, considering $f_s\colon x \mapsto  \ReLU(x) + s(\ReLU(-x) - \ReLU(x) + x) = \ReLU(x)$, we have $\backprop_0[f_s](0) = s$.}

\paragraph{What does backpropagation compute?} A popular thinking is that the impact of $s = \ReLU'(0)$ should be limited as it concerns the value at a single point. This happens to be exact in theory but for surprisingly complex reasons related to Whitney stratification (see Section \ref{sec:mathAutoDiff} and references therein):

---\quad For a given neural network architecture, $\backprop_s$ outputs a gradient almost everywhere, independently of the choice of $s \in \RR$, see \cite{bolte2020conservative} and \cite{berner2019towards} for a detailed treatment of $\ReLU$ networks.

---\quad Under proper randomization of the initialization and the step-size of SGD, with probability $1$ the value of $s$ has no impact during training with $\backprop_s$ as a gradient surrogate, see \cite{bolte2020mathematical,bianchi2020convergence}.

In particular, the set of network parameters such that $\backprop_0 \neq \backprop_1$ is negligible and the vast majority of SGD sequences produced by  training neural networks are not impacted by changing the value of $s = \ReLU'(0)$. 
These results should in principle close the question about the role of $\ReLU'(0)$. Yet, this is not what we observe in practice with the default settings of usual libraries.

\begin{figure}
\centering
    
      \begin{center}
        \includegraphics[width=.3\textwidth]{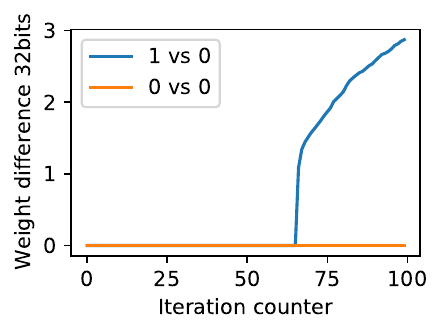} \quad \includegraphics[width=.33\textwidth]{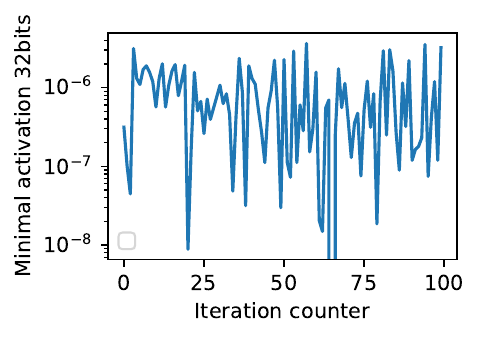} \quad
        \includegraphics[width=.3\textwidth]{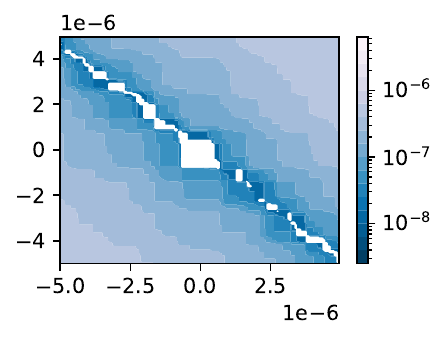} 
      \end{center}
    \caption{Left: Difference between network parameters ($L^1$ norm), 100 iterations within an epoch. ``0 vs 0''  indicates $\|\theta_{k,0} - \tilde{\theta}_{k,0}\|_1$ where $\tilde{\theta}_{k,0}$ is a second run for sanity check, ``0 vs 1'' indicates $\|\theta_{k,0} - \theta_{k,1}\|_1$. Center: minimal absolute activation of the hidden layers within the $k$-th mini-batch, before $\ReLU$. At iteration 65, the jump on the left coincides with the drop in the center, and an exact evaluation of $\ReLU'(0)$. Right: illustration of the bifurcation zone at iteration $k=65$ in a randomly generated network weight parameter plane (iteration 65 in the center). The quantity represented is the absolute value of the neuron of the first hidden layer which is found to be exactly zero within the mini-batch before application of $\ReLU$ (exact zeros are represented in white).}
    \label{fig:first_xp}
\end{figure}

\paragraph{A surprising experiment and the bifurcation zone:}
An empirical observation on MNIST triggered our investigations: consider a fully connected $\ReLU$ network, and let $(\theta_{k,0})_{k \in \NN}$ and $(\theta_{k,1})_{k \in \NN}$ be two training weights sequences obtained using SGD, with the same random initialization and the same mini-batch sequence but choosing respectively $s = 0$ and $s = 1$ in PyTorch \cite{NEURIPS2019_9015}. As depicted in Figure~\ref{fig:first_xp} (left), the two sequences differ. A closer look shows that the sudden divergence is related to what we call the {\em bifurcation zone}, i.e., the set $S_{0,1}$ of weights such that $\backprop_0 \neq \backprop_1$. As mentioned previously this contradicts theory which predicts that the bifurcation zone is met with null probability during training. This contradiction is due to the precision of floating point operations and, to a lesser extent, to the size of deep learning problems. 
Indeed, rounding schemes used for inexact arithmetics over the reals (which set to zero all values below a certain threshold), may ``thicken'' negligible sets. This is precisely what happens in our experiments (Figure \ref{fig:first_xp}). 

\paragraph{The role of numerical precision:} Contrary to numerical linear algebra libraries such as \texttt{numpy}, which operates by default under 64 bits precision, the default choice in PyTorch is 32 bits precision (as in TensorFlow or Jax). We thus modulated machine precision to evaluate the importance of the bifurcation zone in Section \ref{sec:prelimMNIST}. In 32 bits, we observed that empirically the zone occupies from about 10\% to 90\% of the network weight space. It becomes invisible in 64 bits precision even for quite large architectures, while, in 16 bits, it systematically fills up the whole space. Although numerical precision is the primary cause of the apparition of the zone, we identify other factors such as network size, sample size. Let us mention that low precision neural network training is a topic of active research \cite{vanhoucke2011improving,hwang2014fixed,courbariaux2015training,gupta2015deep}, see also \cite{rodriguez2018lower} for an overview. Our investigations are complementary as we focus on the interplay between nonsmoothness and numerical precision.

\paragraph{Impact on learning:} The next natural question is to measure the impact of the choice of $s = \ReLU'(0)$ in machine learning terms. In Section \ref{sec:consequencesLearning}, we conduct extensive experiments combining different architectures (fully connected, VGG, ResNet), datasets (MNIST, SVHN, CIFAR10, ImageNet) and other learning factors (Adam  optimizer, batch normalization, dropout). In 32 bits numerical precision (default in PyTorch or Tensorflow), we consistently observe an effect of choosing $s \neq 0$. We observe a significant decrease in terms of test accuracy as $|s|$ increases; this can be explained by chaotic oscillatory behaviors induced during training. In some cases gradients even explode and learning cannot be achieved. The sensitivity to this effect highly depends on the problem at hand, in particular, on the network structure and the dataset. On the other hand the choice $s = 0$ seems to be the most  stable. We also observe that both batch normalization \cite{ioffe2015batch} and---to a  lesser degree---the Adam  optimizer \cite{kingma2015adam} considerably mitigate this effect. All our experiments are done using PyTorch \cite{NEURIPS2019_9015}; we provide the code to generate all figures presented in this manuscript.

One important message is that, even if the default choice $s=0$ seems to be the most stable, our experiments show a counter-intuitive phenomenon that illustrates the interplay between numerical precision and nonsmoothness, and calls for caution when learning nonsmooth networks.

\paragraph{Outline of the paper:} In Section \ref{sec:mathAutoDiff} we recall elements of nonsmooth algorithmic differentiation which are key to understand the mathematics underlying our experiments. Most results were published in \cite{bolte2020conservative,bolte2020mathematical}; we provide more detailed pointers to this literature in Appendix \ref{sec:additionalMath}. In Section \ref{sec:prelimMNIST} we describe investigations of the bifurcation zone and factors influencing its importance using fully connected networks on the MNIST dataset. Neural network training experiments are detailed in Section \ref{sec:consequencesLearning} with further experimental details and additional experiments reported in Appendix \ref{sec:additionalExpe}.

\section{On the mathematics of backpropagation for ReLU networks}
\label{sec:mathAutoDiff}
This section recalls recent advances on the mathematical properties of backpropagation, with in particular the almost sure absence of impact of $\ReLU'(0)$ on the learning phase (assuming exact arithmetic over the reals). The main mathematical tools are conservative fields developed in \cite{bolte2020conservative}; we provide a simplified overview which is applicable to a wide class of neural networks.
\subsection{Empirical risk minimization and backpropagation}
Given a training set $\{(x_i,y_i)\}_{i = 1 \ldots N}$, the supervised training of a neural network $f$ consists in minimizing the empirical risk: 
\begin{align}
    \min_{\theta \in \RR^P} \quad J(\theta) :=  \frac{1}{N} \sum_{i=1}^N \ell(f(x_i, \theta), y_i)
    \label{eq:NNtraining}
\end{align}
where $\theta \in \RR^P$ are the network's weight parameters and $\ell$ is a loss function. The problem can be rewritten abstractly, for each $i = 1, \ldots, N$ and $\theta \in \RR^P$, $\ell(f(x_i, \theta), y_i) = l_i(\theta)$ 
where the function $l_i \colon \RR^P \to \RR$ is a composition of the form
\begin{align}
    l_i = g_{i,M} \circ g_{i,M-1} \circ \ldots \circ g_{i,1}
    \label{eq:composition}
\end{align}
where for each $j = 1, \ldots, M$, the function $g_{i,j}$ is locally Lipschitz with appropriate input and output dimensions. A concrete example of what the functions $g_{i,j}$ look like is given in Appendix~\ref{sec:gijFullyConnected} in the special case of fully connected  ReLU networks. Furthermore, we associate with each $g_{i,j}$ a generalized Jacobian $J_{i,j}$ which is such that $J_{i,j} (w) = \mathrm{Jac}_{g_{i,j}} (w)$ whenever $g_{i,j}$ is differentiable at $w$ and $\mathrm{Jac}$ denotes the usual Jacobian. The value of $J_{i,j}$ at the nondifferentiability loci of $g_{i,j}$ can be arbitrary for the moment. The backpropagation algorithm is an automatized implementation of the rules of differential calculus: for each $i = 1, \ldots, N$, we have
\begin{align}
    \backprop\ l_i (\theta) = J_{i,M}\left( g_{i, M-1} \circ \ldots \circ g_{i,1} (\theta) \right) \times J_{i,M-1}\left( g_{i, M-2} \circ \ldots \circ g_{i,1} (\theta) \right) \times \ldots \times J_{i,1}(\theta).
    \label{eq:defBackprop}
\end{align}
Famous autograd libraries such as PyTorch \cite{NEURIPS2019_9015} or TensorFlow \cite{tensorflow2015-whitepaper} implement dictionaries of functions $g$ with their corresponding generalized Jacobians $J$, as well as efficient numerical implementation of the quantities defined in \eqref{eq:defBackprop}.

\subsection{ReLU networks training}

Our main example is based on the function $\ReLU \colon \RR \to \RR$ defined by $\ReLU(x)=\max\{x,0\}$. It is differentiable save at the origin and satisfies $\ReLU'(x) = 0$ for $x < 0$ and $\ReLU'(x) = 1$ for $x > 0$. The value of the derivative at $x=0$ could be arbitrary in $[0,1]$ as we have $\partial \ReLU(0) = [0,1]$, where $\partial$ denotes the subgradient from convex analysis. Let us insist on the fact that any value within $[0,1]$ has a variational meaning. For example PyTorch and TensorFlow use $\ReLU'(0) = 0$. 

Throughout the paper, and following the lines of \cite{bolte2020mathematical}, we say that a function $g:\RR^p \to \RR^q$ is \emph{elementary \logexp} if it can be described by a finite compositional expression involving the basic operations $+, -, \times, /$ as well as the exponential and logarithm functions, inside their domains of definition. Examples include the logistic loss $\log(1+e^{-x})$ on $\RR$, the multivariate Gaussian density $(2 \pi)^{-K/2} \exp(-\sum_{k=1}^K x_k^2/2)$ on $\RR^K$, and the softmax function $\bigl(e^{x_i}/\sum_{k=1}^{K} e^{x_k}\bigr)_{1 \leq i \leq K}$ on~$\RR^K$. The expressions $x^3/x^2$ and $\exp(-1/x^2)$ do not fit this definition because evaluation at $x=0$ cannot be defined by the formula. Roughly speaking a computer evaluating an elementary $\log$-$\exp$ expression should not output any NaN error for any input. 

\begin{definition}[ReLU network training]\label{def:reluNetwork}{\rm
    Assume that in \eqref{eq:NNtraining}, the function $\ell$ is elementary \logexp, the network $f$ has an arbitrary structure and the functions  $g_{i,j}$ in \eqref{eq:composition} are either elementary {\logexp} or consist in applying the $\ReLU$ function to some coordinates of their input. We then call the problem in \eqref{eq:NNtraining} a \emph{$\ReLU$ network training problem}. Furthermore, for any $s \in \RR$, we denote by $\backprop_s$ the quantity defined in \eqref{eq:defBackprop} when $\ReLU'(0) = s$ for all $\ReLU$ functions involved in the composition \eqref{eq:composition}.}
\end{definition}

\paragraph{Other nonsmooth activation functions: } The $\ReLU$ operation actually allows to express many other types of nonsmoothness such as absolute values, maxima, minima, quantiles ($\mathrm{med}$ for median) or soft-thresholding ($\mathrm{st}$). For any $x,y,z \in \RR$, we indeed have $|x|=\ReLU(2x)-x$, $2\max(x,y)=(|x-y| +x + y)$, $\min(x,y) = -\max(-x,-y)$,
    $\mathrm{med}(x,y,z) = \min(\max(x,y), \max(x,z), \max(y,z))$, $\,\mathrm{st}(x) = \ReLU(x - 1) - \ReLU(-x-1)$.


Definition \ref{def:reluNetwork} is thus much more general than it may seem since it allows, for example, to express  convolutional neural networks with max-pooling layers such as VGG or ResNet architectures which correspond to the models considered in the experimental section (although we do not re-program pooling using $\ReLU$). The following theorem is due to \cite{bolte2020conservative}, with an alternative version in \cite{bolte2020mathematical}.
\begin{theorem}[Backprop returns a gradient a.e.]
        Consider a $\ReLU$ network training problem \eqref{eq:NNtraining} as in Definition \ref{def:reluNetwork} and $T \geq 1$. Define $S \subset \RR^P$ as the complement of the set
        \begin{align*}
            \left\{\theta \in \RR^P,\, l_i \text{ differentiable at } \theta,\, \backprop_s l_i(\theta) = \nabla l_i(\theta), \qquad \forall i \in  \{1, \ldots, N\}, \, s \in [-T,T]\right\}.
        \end{align*}
        Then $S$ is contained in a finite union of embedded differentiable manifolds of dimension at most $P-1$ (and in particular has Lebesgue measure zero).
        \label{th:mainBP}
\end{theorem}
Although this theorem looks natural, this is a nontrivial result about the backpropagation algorithm that led to the introduction of conservative fields \cite{bolte2020conservative}. It implies that all choices for $s = \ReLU'(0)$ in $[0,1] = \partial \ReLU(0)$ are essentially equivalent modulo a negligible set $S$. Perhaps more surprisingly, $s$ can be chosen arbitrarily in $\RR$ without breaking this essential property of $\backprop$. The set $S$ is called the \emph{bifurcation zone} throughout the manuscript. For $\ReLU$ network training problems, the bifurcation zone is a Lebesgue zero set and is actually contained locally in a finite union of smooth objects of dimension strictly smaller than the ambient dimension. This geometric result reveals a surprising link between backpropagation and Whitney stratifications, as described in \cite{bolte2020conservative,bolte2020mathematical}. In any case the bifurcation zone is completely negligible. Note that the same result holds if we allow each different call to the $\ReLU$ function to use different values for $\ReLU'(0)$.

\subsection{ReLU network training with SGD}
Let $(B_k)_{k \in \NN}$ denote a sequence of mini-batches with sizes $|B_k|\ \subset \{1,\ldots, N\}$ for all $k$ and $\alpha_k > 0$ the learning rate.
Given  initial weights $\theta_0 \in \RR^P$ and any parameter $s \in \RR$, the SGD training procedure of $f$ consists in applying the recursion
\begin{align}
    \theta_{k+1,s} = \theta_{k,s} - \gamma \frac{\alpha_k}{|B_k|} \sum_{n \in B_k} \backprop_s[\ell(f(x_n, \theta_{k,s}), y_n)],
    \label{eq:SGDbackprop}
\end{align}
where $\gamma > 0$ is a step-size parameter. Note that we explicitly wrote the dependency of the sequence in $s = \ReLU'(0)$. According to Theorem \ref{th:mainBP} if the initialization $\theta_0$ is chosen randomly, say, uniformly in a ball, a hypercube, or with iid Gaussian entries, then with probability $1$, $\theta_0$ does not fall in the bifurcation zone $S$. Intuitively, since $S$ is negligible, the odds that one of the iterates produced by the algorithm fall on $S$ are very low. As a consequence, varying $s$ in the recursion \eqref{eq:SGDbackprop} does not modify the sequence. This rationale is actually true for almost all values of $\gamma$.  This provides a rigorous  statement of the idea that the choice of $\ReLU'(0)$ ``does not affect'' neural network training. The following result is based on  arguments developed in \cite{bolte2020mathematical}, see also \cite{bianchi2020convergence} in a probabilistic context.

\begin{theorem}[$\ReLU'(0)$ does not impact training with probability 1]
    Consider a $\ReLU$ network training problem as in Definition \ref{def:reluNetwork}. Let $(B_k)_{k \in \NN}$ be a sequence of mini-batches with $|B_k|\ \subset \{1,\ldots, N\}$ for all $k$ and $\alpha_k > 0$  the associated learning rate parameter. Choose $\theta_0$ uniformly at random in a hypercube and $\gamma$ uniformly in a bounded interval $I \subset \RR_+$. Let $s \in \RR$, set $\theta_{0,s} = \theta_0$, and consider the recursion given in \eqref{eq:SGDbackprop}. Then, with probability one, for all $k \in \NN$, $\theta_{k,s} = \theta_{k,0}$.
    \label{th:noInfluenceOnSequence}
\end{theorem}


\section{Surprising experiments on a simple feedforward network}
\label{sec:prelimMNIST}
\subsection{$\ReLU'(0)$ has an impact}

\label{sec:choice_of_relu}
Even though the $\ReLU$ activation function is non-differentiable at 0, autograd libraries such as PyTorch \cite{NEURIPS2019_9015} or TensorFlow \cite{tensorflow2015-whitepaper} implement its derivative with $\ReLU'(0) = 0$.
What happens if one chooses $\ReLU'(0) = 1$? The popular answer to this question is that it should have no effect. Theorems \ref{th:mainBP} and \ref{th:noInfluenceOnSequence} provide a formal justification which is far from being trivial.

\paragraph{A 32 bits MNIST experiment} We ran a simple experiment to confirm this answer. We initialized two fully connected neural networks $f_0$ and $f_1$ of size $784 \times 2000 \times 128 \times 10$ with the same weights $\theta_{0,0} = \theta_{0,1} \in \RR^P$ which are chosen at random.
Using the MNIST dataset \cite{lecun1998gradient}, we trained $f_0$ and $f_1$ with the same sequence of mini-batches $(B_k)_{k \in \NN}$ (minibatch size 128), using the recursion in \eqref{eq:SGDbackprop} for $s = 0$ and $s=1$ and with a fixed $\alpha_k = 1$, and $\gamma$ chosen uniformly at random in  $[0.01,0.012]$. At each iteration $k$, we computed the sum $\|\theta_{k,0} - \theta_{k,1}\|_1$ of the absolute differences between the coordinates of $\theta_{k,0}$ and $\theta_{k,1}$. As a sanity check, we actually computed $\theta_{k,0}$ a second time, denoting this by $\tilde{\theta}_{k,0}$, using a third network to control for sources of divergence in our implementation. Results are reported in Figure~\ref{fig:first_xp}. The experiment was run using PyTorch \cite{NEURIPS2019_9015} on a CPU. 

\paragraph{ReLU'(0) has an impact} First we observe no difference between $\theta_{k,0}$ and $\tilde{\theta}_{k,0}$, which shows that we have controlled all possible sources of divergence in PyTorch. Second,  while no differences between $\theta_{0,0}$ and $\theta_{0,1}$ is expected (Theorem \ref{th:noInfluenceOnSequence}), we observe a sudden deviation of $\|\theta_{k,0} - \theta_{k,1}\|_1$ at iteration 65 which then increases in a smooth way. The deviation is sudden as the norm is exactly zero before iteration 65 and jumps above one after. Therefore this cannot be explained by an accumulation of small rounding errors throughout iterations, as this would produce a smooth divergence starting at the first iteration. So this suggests that there is a special event at iteration 65.

The center part of Figure \ref{fig:first_xp} displays the minimal absolute value of neurons of the first hidden layer evaluated on the current mini-batch, before the application of $\ReLU$. It turns out that at iteration 65, this minimal value is exactly $0$, resulting in a drop in the center of Figure \ref{fig:first_xp}. This means that the divergence is actually due to an iterate of the sequence falling exactly on the bifurcation zone. According to Theorem \ref{th:noInfluenceOnSequence}, this event is so exceptional that it should never been seen.

\paragraph{Practice and Theory: Numerical precision vs Genericity} This contradiction can be solved as follows: the minimal absolute value in Figure \ref{fig:first_xp} oscillates between $10^{-6}$ and $10^{-8}$ which is roughly the value of machine precision in 32 bits float arithmetic. This machine precision value is of the order $10^{-16}$ in 64 bits floating arithmetic which is orders of magnitude smaller than the typical value represented in Figure~\ref{fig:first_xp}. And indeed, performing the same experiment in 64 bits precision, the divergence of  $\|\theta_{k,0} - \theta_{k,1}\|_1$ disappears and the algorithm can actually be run for many epochs without any divergence between the two sequences. This is represented in Figure \ref{fig:first_xp_64bits} in Appendix \ref{sec:firstExp64BitsPrec}. We also report similar results using $\ReLU6$ \cite{howard2017mobilenets} in place of $\ReLU$ on a similar network.

\subsection{Relative volume of the bifurcation zone and relative gradient variation}
\label{sec:xp_proc}


The previous experiment suggests that  mini-batch SGD algorithm \eqref{eq:SGDbackprop} crossed the bifurcation zone:
\begin{align}
    S_{01} = \{\theta \in \RR^P :\ \exists i \in \{1,\ldots, N\},\  \backprop_0[l_i](\theta) \neq \backprop_1[l_i](\theta)\}\subset S.
    \label{eq:S01}
\end{align}
This unlikely phenomenon is due to finite numerical precision arithmetic which thickens the negligible set $S_{01}$ in proportion to the machine precision threshold. This is illustrated on the right of Figure~\ref{fig:first_xp}, which represents the bifurcation zone at iteration 65 in a randomly generated hyperplane (uniformly among 2 dimensional hyperplanes) centered around  the weight parameters which generated the bifurcation in the previous experiment (evaluation on the same mini-batch). The white area corresponds to some entries being exactly zero, i.e., below the 32 bits machine precision threshold, before application of $\ReLU$. On the other hand, in 64 bits precision, the same representation is much smoother and does not contain exact zeros (see Figure \ref{fig:first_xp_64bits} in Appendix~\ref{sec:firstExp64BitsPrec}). This confirms that the level of floating point arithmetic precision explains the observed divergence. We now estimate the relative volume of this bifurcation zone by Monte Carlo sampling (see Appendix \ref{sec:montecarlo} for details). All experiments are performed using PyTorch \cite{NEURIPS2019_9015} on GPU.

\paragraph{Experimental procedure -- weight randomization:}
We randomly generate a set of parameters $\{\theta_j\}_{j=1}^M$, with $M=1000$, for a fully connected network architecture $f$ composed of $L$ hidden layers. Given two consecutive layers, respectively composed of $m$ and $m'$ neurons, the weights of the corresponding affine transform are drawn independently, uniformly in $[-\alpha, \alpha]$ where $\alpha=\sqrt{6}/\sqrt{m}$. This is the default weight initialization scheme in PyTorch (Kaiming-Uniform \cite{he2015delving}). Given this sample of parameters, iterating on the whole MNIST dataset, we approximate the proportion of $\theta_j$ for which $ \backprop_0(l_i)(\theta_j) \neq \backprop_1(l_i)(\theta_j)$ for some $i$, for different networks and under different conditions (see Appendix \ref{sec:montecarlo} for details).

\paragraph{Impact of the floating-point precision:}

Using a fixed architecture of three hidden layers of 256 neurons each, we empirically measured the relative volume of $S_{01}$ using the above experimental procedure, varying the numerical precision.
Table \ref{table:precision} reports the obtained estimates.
As shown in Table \ref{table:precision}, line 1, at 16 bits floating-point precision, all drawn weights  $\{\theta_i\}_{i=1}^M$ fall within $S_{01}$. 
In sharp contrast, when using a 64 bits precision, none of the sampled weights belong to $S_{01}$. 
This proportion is 40\% in 32 bits precision. 
For the rare impacted mini-batches (Table \ref{table:precision} line 2), the relative change in norm is above a factor 20, higher in 16 bits precision (Table \ref{table:precision} line 3).
These results confirm that the floating-point arithmetic precision is key in explaining the impact of $\ReLU'(0)$ during backpropagation, impacting both frequency and magnitude of the differences.

\begin{table}[ht]
\begin{center}
 \begin{tabular}{||l || c c c||} 
 \hline
 Floating-point precision & 16 bits & 32 bits & 64 bits \\
 \hline
 \hline
 Proportion of $\{\theta_i\}_{i=1}^M$ in $S$ (CI $\pm$ 5\%) & 100\% & 40\%  & 0\% \\  
 \hline
 Proportion of impacted mini-batches  (CI $\pm$ 13\%) & 0.05\% & 0.0002\% & 0\% \\  
 Relative $L^2$ difference for impacted mini-batches &&&  \\
 (1st quartile, median, 3rd quartile)& (98,\,117,\,137) & (19,\,25,\,47) & (0,\,0,\,0)\\
 \hline

\end{tabular}
\end{center}
 \caption{Impact of $S$ according to the floating-point precision on a fully connected neural network ($784 \times 256 \times 256 \times 256 \times 10$) on MNIST. First line: proportion of drawn weights $\theta_i$ such that at least one mini-batch results in difference between $\backprop_0$ and $\backprop_1$. CI stands for \emph{Confidence Interval} with $5 \%$ risk (Hoeffding CI, see Appendix~\ref{sec:montecarlo}). Second line: overall proportion of mini-batch-weight vector pairs causing a difference between $\backprop_0$ and $\backprop_1$ (McDiarmid CI, see Appendix~\ref{sec:montecarlo}). Third line: distribution of $\|\backprop_0 - \backprop_1\|_2 / \|\backprop_0\|_2$ for the affected mini-batch-weight vector pairs.}
 \label{table:precision}
 \vskip-.2in
\end{table}

\begin{figure}[t]
  \begin{center}
    \includegraphics[width=\textwidth]{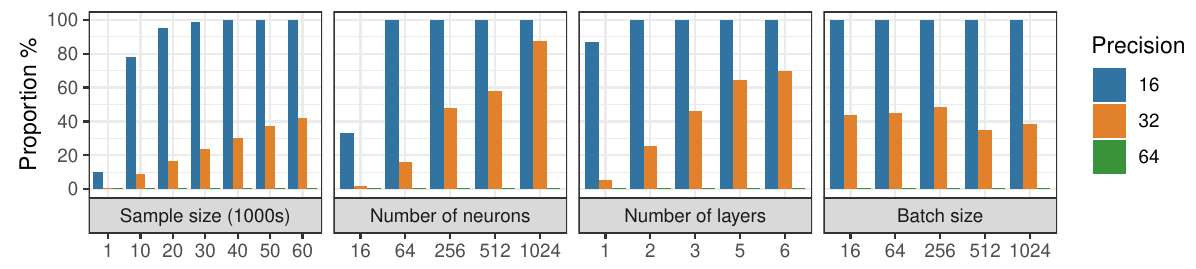}
    \end{center}
  \caption{Influence of different size parameters on the proportion of drawn weight vector $\theta_i$ such that at least one mini-batch results in difference between $\backprop_0$ and $\backprop_1$ on MNIST dataset. Confidence interval at risk level $5\%$ results in a variation of $\pm 5 \%$ for all represented proportions. First: 4 hidden layers with 256 neurons and batch size 256, varying the size of the training set. Second: 3 hidden layer network with mini-batch size 256, varying the number of neuron per layer. Third: 256 neurons per layer and mini-batch size 256, varying the number of layers. Fourth: 3 hidden layers with 256 neurons per layer, varying mini-batch size.}
\label{fig:network_size}
\end{figure}

\paragraph{Impact of sample and mini-batch size:}
Given a training set $\{(x_n,y_n)\}_{n = 1 \ldots N}$ and a random variable $\theta \in \RR^P$ with distribution $\PP_\theta$, we estimate the probability that $\theta \in S_{01} \subset S$. Intuitively, this probability should increase with sample size $N$. We perform this estimation for a fixed architecture of 4 hidden layers composed of 256 neurons each while varying the sample size. Results are reported in Figure~\ref{fig:network_size}. For both the 16 and the 32 bits floating-point precisions, our estimation indeed increases with the sample size while we do not find any sampled weights in $S_{01}$ in 64 bits precision. We also found that the influence of mini-batch size is not significative.

\paragraph{Impact of network size:}
To evaluate the impact of the network size, we carried out a similar Monte Carlo experiment, varying the depth and width of the network. Firstly, we fixed the number of hidden layers to 3. Following the same experimental procedure, we empirically estimated the relative volume of $S_{01}$, varying the width of the hidden layers. The results, reported in Figure \ref{fig:network_size}, show that increasing the number of neurons by layer increases the probability to find a random $\theta \sim \PP_\theta$ in $S_{01}$ for both 16 and 32 floating-point precision. In 64 bits, even with the largest width tested (1024 neurons), no sampled weight parameter is found in $S_{01}$.
Similarly we repeated the experiment varying the network depth and fixing, this time, the width of the layers to 256 neurons. Anew, the results, reported in Figure \ref{fig:network_size}, show that increasing the network depth increases the probability that $\theta \sim \PP_\theta$ belongs to $S_{01}$ for both 16 and 32 bits floating-point precision while this probability is zero in 64 bits precision. This shows that the size of the network, both in terms of number of layers and size of layers is positively related to the effect of the choice of $s$ in backpropagation. On the other hand, the fact that neither the network depth, width, or the number of samples impact the 64 bits case suggests that, within our framework, numerical precision is the primary factor of deviation.

\section{Consequences for learning}
\label{sec:consequencesLearning}

\subsection{Benchmarks and implementation}

\paragraph{Datasets and networks}
We further investigate the effect of the phenomenon described in Section~\ref{sec:prelimMNIST} in terms of learning using the CIFAR10 dataset \cite{krizhevsky2010cifar} and the VGG11 architecture \cite{simonyan15very}. To confirm our findings in alternative settings, we also use the MNIST \cite{lecun1998gradient}, SVHN \cite{netzer2011reading} and ImageNet \cite{deng2009imagenet} datasets, fully connected networks (3 hidden layers of size 2048), and the ResNet18 and ResNet50 architectures \cite{he2016deep}. Additional details on the different architectures and datasets are found in Appendix \ref{sec:benchmarksAndArchitectures}. By default, unless stated otherwise, we use the SGD optimizer. We also investigated the effect of batch normalization \cite{ioffe2015batch}, dropout \cite{srivastava2014dropout}, the Adam optimizer \cite{kingma2015adam} as well as numerical precision. All the experiments in this section are run in PyTorch \cite{NEURIPS2019_9015} on GPU. For each training experiment presented in this section (except ImageNet experiments), we use the \texttt{optuna} library \cite{akiba2019optuna} to tune learning rates for each experimental condition; see also Appendix \ref{sec:additionalMNIST}. 

\subsection{Effect on training and test errors}
\label{sec:numMainEffect}
\begin{figure}
\centering
  \begin{center}
    \includegraphics[width=.52
    \textwidth]{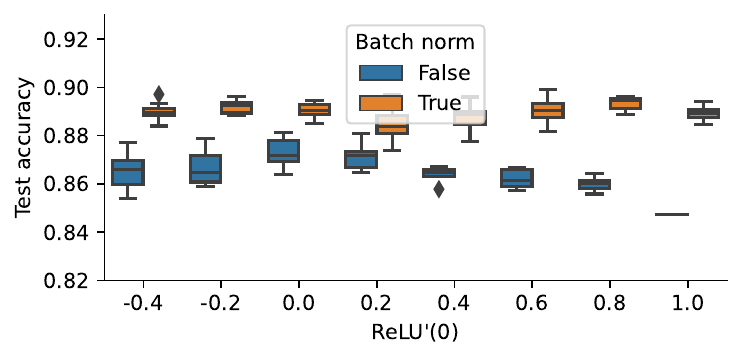} \quad
    \includegraphics[width=.43\textwidth]{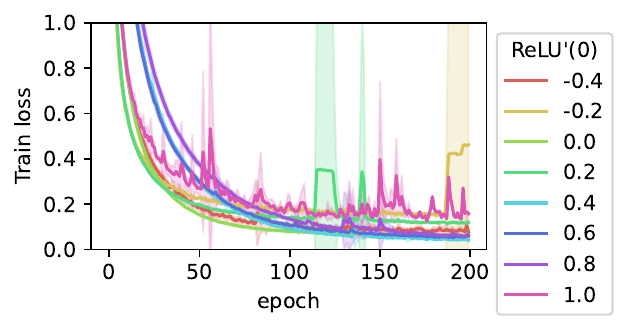}
  \end{center}
  \caption{Training a VGG11 network on CIFAR10 with SGD. Left: test accuracy with and without batch normalization. Right: training loss without batchnorm. For each experiment we performed 10 random initializations, depicted by the boxplots on the left and the filled contours on the right (standard deviation).}
\label{fig:cifarVGG}
\end{figure}

We first consider training a VGG11 architecture on CIFAR10 using the SGD optimizer. For different values of $\ReLU'(0)$, we train this network ten times with random initializations under 32 bits arithmetic precision. The results are depicted in Figure \ref{fig:cifarVGG}. Without batch normalization, varying the value of $\ReLU'(0)$ beyond a magnitude of $0.1$ has a 
detrimental effect on the test accuracy, resulting in a concave shaped curve with maximum around $\ReLU'(0) = 0$. On the other hand, the decrease of the training loss with the number of epochs suggests that choosing $\ReLU'(0) \neq 0$  induces jiggling behaviors with possible sudden jumps during training. Note that the choice $\ReLU'(0) = 0$ leads to a smooth decrease and that the magnitude of the jumps for other values is related to the magnitude of the chosen value. This is consistent with the interpretation that changing the value of  $\ReLU'(0)$ has an excitatory effect on training. We observed qualitatively similar behaviors for a fully connected network on MNIST and a ResNet18 on CIFAR10 (Appendix \ref{sec:additionalExpe}). Sensitivity to the magnitude of $\ReLU'(0)$ depends on the network architecture: our fully connected network on MNIST is less sensitive to this value than VGG11 and ResNet18. The latter shows a very high sensitivity since for values above $0.2$, training becomes very unstable and almost impossible. 

These experiments complement the preliminary results obtained in Section \ref{sec:prelimMNIST}. In particular choosing different values for $\ReLU'(0)$ has an effect, it induces a chaotic behavior during training which affects test accuracy. The default choice $\ReLU'(0) = 0$ seems to provide the best performances. 

To conclude, we conducted four training experiments for ResNet50 on the ImageNet dataset \cite{deng2009imagenet} using the SGD optimizer. These were conducted with fixed learning rate, contrary to results reported above. We observe that switching from $\ReLU'(0) = 0$ to $1$ results in a massive drop from around $75\%$ to $63\%$ or $55\%$ for two runs.

\subsection{Mitigating factors: numerical precision, batch-normalization and Adam }
\label{sec:numBatchNorm}
We analyze below the combined effects of the variations of $\ReLU'(0)$ with numerical precision or classical reconditioning methods: Adam, batch-normalization and dropout.
\paragraph{Batch-normalization:} As represented in Figure \ref{fig:cifarVGG}, batch normalization \cite{ioffe2015batch} not only allows to attain higher test accuracy, but it also completely filters out the effect of the choice of the value of $\ReLU'(0)$, resulting in a flat shaped curve for test accuracy. This is consistent with what we observed on the training loss (Figure \ref{fig:trainWithBatchNorm} in Appendix \ref{sec:additionalVGG11}) for which different choices of $\ReLU'(0)$ lead to indistinguishable training loss curves. This experiment suggests that batch normalization has a significant impact in reducing the effect of the choice of $\ReLU'(0)$. This observation was confirmed with a very similar behavior on the MNIST dataset with a fully connected network (Figure~\ref{fig:MNISTTest} in Appendix~\ref{sec:additionalMNIST}). We could observe a similar effect on CIFAR 10 using a ResNet18 architecture (see Appendix \ref{sec:additionalResnet}), however in this case the value of $\ReLU'(0)$ still has a significative impact on test error, the ResNet18 architecture being much more sensitive.

\begin{figure}
\centering
    \includegraphics[width=.48\textwidth]{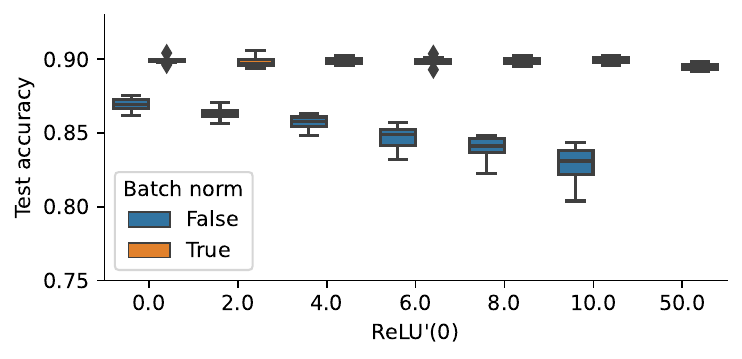}
    \quad
    \includegraphics[width=.48\textwidth]{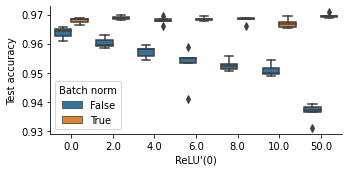} 
  \caption{Similar to Fig.\ref{fig:cifarVGG}, with VGG11 and Adam optimizer, on CIFAR 10 (left) and SVHN (right).}
\label{fig:SVHNVGG}
\end{figure}

\paragraph{Using the Adam optimizer:} The Adam optimizer \cite{kingma2015adam} is among the most popular algorithms for neural network training; it combines adaptive step-size strategies with momentum. Adaptive step-size acts as diagonal preconditioners for gradient steps \cite{kingma2015adam,duchi2011adaptive} and therefore can be seen as having an effect on the loss landscape of neural network training problems. We trained a VGG11 network on both CIFAR 10 and SVHN using the Adam optimizer. The results are presented in Figure \ref{fig:SVHNVGG}. We observe a qualitatively similar behavior as in the experiments of Section~\ref{sec:numMainEffect} but a much lower sensitivity to the magnitude of $\ReLU'(0)$. In other words, the use of the Adam optimizer mitigates the effect of this choice, both in terms of test errors and by buffering the magnitude of the sometimes chaotic effect induced on training loss optimization (Figure \ref{fig:cifarVGGAdam} in Appendix \ref{sec:additionalVGG11}).

\paragraph{Increasing numerical precision:} As shown in Appendix \ref{sec:additionalVGG11}, using 64 bits floating precision on VGG11 with CIFAR10 cancels out the effect of $\ReLU'(0) = 1$, in coherence with Section \ref{sec:prelimMNIST}. More specifically $\ReLU'(0) = 1$ in 64 bits precision obtains similar performances as $\ReLU'(0) = 0$ in 32 bits precision. Furthermore, the numerical precision has barely any effect when $\ReLU'(0) = 0$. Finally, we remark that in 16 bits with $\ReLU'(0)=1$, training is extremely unstable so that we were not able to train the network in this setting.

\paragraph{Combination with dropout:} Dropout \cite{srivastava2014dropout} is another algorithmic process to regularize deep neural networks. We investigated its  effect when combined with varying choices of $\ReLU'(0)$. We used a fully connected neural network trained on the MNIST dataset with different values of dropout probability. The results are reported in Figure \ref{fig:MNISTDropout} in Appendix \ref{sec:additionalMNIST}. We did not observe any joint effect of dropout probability and magnitude of $\ReLU'(0)$ in this context.

\subsection{Back to the bifurcation zone: more neural nets and the effects of batch-norm}
\begin{figure}
\centering
  \begin{center}
    \includegraphics[width=.3\textwidth]{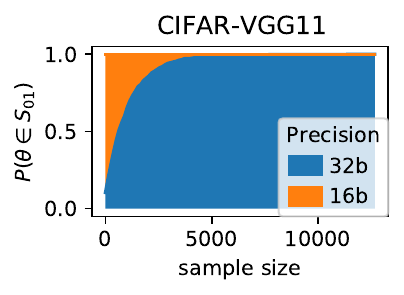} \quad
    \includegraphics[width=.3\textwidth]{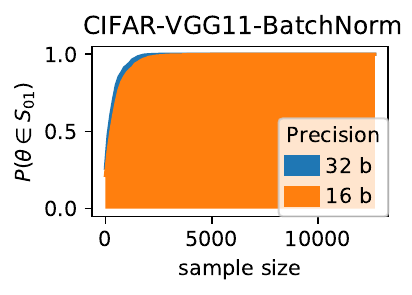} \quad
    \includegraphics[width=.3\textwidth]{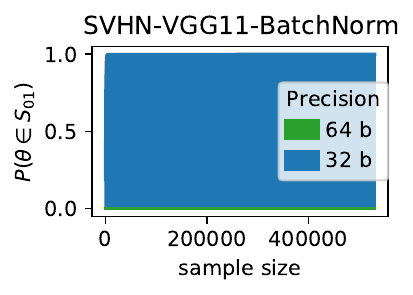}
  \end{center}
  \caption{Monte Carlo estimation of relative volume on CIFAR10 and SVHN with VGG11 network}
\label{fig:SVHNVGGVolume}
\end{figure}

The training experiments are complemented by a similar Monte Carlo estimation of the relative volume of the bifurcation zone as performed in Section \ref{sec:prelimMNIST} (same experimental setting). To avoid random outputs we force the GPU to compute convolutions deterministically. Examples of results are given in Figure~\ref{fig:SVHNVGGVolume}. Additional results on fully connected network with MNIST and ResNet18 on CIFAR10 are shown in Section \ref{sec:additionalExpe}. We consistently observe a high probability of finding an example on the bifurcation zone for large sample sizes. Several comments are in order.

\paragraph{Numerical precision:} Numerical precision is the major factor in the thickening of the bifurcation zone. In comparison to 32 bits experiments, 16 bits precision dramatically increases its relative importance. We also considered 64 bits precision on SVHN, a rather large dataset. Due to the computational cost, we only drew 40 random weights and observed no bifurcation on any of the terms of the loss whatsoever. This is consistent with the experiments conducted in Section \ref{sec:prelimMNIST} and suggests that, within our framework, 64 bit precision is the main mitigating factor for our observations.

\paragraph{Batch normalization:} In all our relative volume estimation experiments, we observe that batch normalization has a very significant effect on the proportion of examples found in the bifurcation zone. In 32 bits precision, the relative size of this zone increases with the addition of batch normalization, similar observations were made in all experiments presented in Appendix~\ref{sec:additionalExpe}. This is a counter-intuitive behavior as we have observed that batch normalization increases test accuracy and mitigates the effect of $\ReLU'(0)$. Similarly in 16 bits precision, the addition of batch normalization seems to actually decrease the size of the bifurcation zone. Batch normalization does not result in the same qualitative effect depending on arithmetic precision. These observations open many questions which  will be the topic of future research.

\subsection{Total number of $\ReLU$ calls during training}
\begin{figure}
\centering
  \begin{center}
    \includegraphics[width=.5\textwidth]{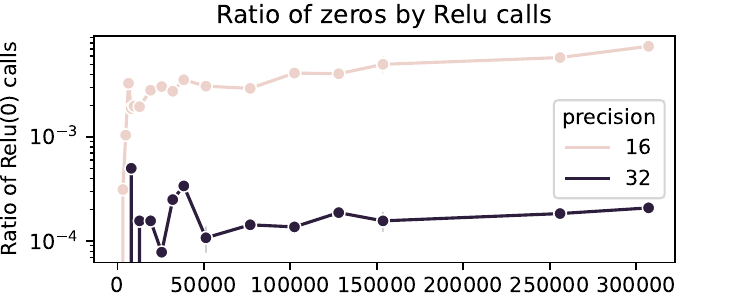} 
  \end{center}
  \caption{Proportion of $\ReLU(0)$ calls by total number of $\ReLU$ calls after 100 epochs for $\ReLU$ networks on MNIST with number of layers in $1,\ldots,6$ and neuron per layers in $16,64,256,512$. \\[-0.7cm]}
\label{fig:propZeroCall}
\end{figure}

We consider the MNIST dataset with a fully connected $\ReLU$ network, varying the number of layers in $1,\ldots,6$ and neuron per layers in $16,64,256,512$. For $16$ and $32$ bits precisions, we perform $100$ epochs and proportion of $\ReLU(0)$ over all $\ReLU$ calls during training. Figure \ref{fig:propZeroCall} suggests that, for a given precision, the number of times the bifurcation zone is met is roughly proportional to the total number of $\ReLU$ calls during training, independently of the architecture used (number of layers, neurons per layers). This suggests that the total number of $\ReLU$ calls during training is an important factor in understanding the bifurcation phenomenon. Broader investigation of this conjecture will be a matter of future work.

\section{Conclusions and future work}
The starting point of our work was to determine if the choice of the value $s = \ReLU'(0)$ affects neural network training. Theory tells that this choice should have negligible effect. Performing a simple learning experiment, we discovered that this is false in the real world and the first purpose of this paper is to account for this empirical evidence. This contradiction between theory and practice is due to finite floating point arithmetic precision while idealized networks  are analyzed theoretically within a model of exact arithmetic on the field of real numbers. 
Owing to the size of deep learning problems, rounding errors due to numerical precision  occur at a relatively high frequency, and virtually all the time for large architectures and datasets under 32 bit arithmetic precision (the default choice for TensorFlow and PyTorch libraries).

Our second goal was to investigate the impact of the choice of $s = \ReLU'(0)$ in machine learning terms. In 32 bits precision it has an effect on test accuracy which seems to be the result of inducing a chaotic behavior in the course of empirical risk minimization. This was observed consistently in all our experiments. However we could not identify a systematic quantitative description of this effect; it highly depends on the dataset at hand, the network structure as well as other learning parameters such as the presence of batch normalization and the use of different optimizers. Our experiments illustrate this diversity. We observe an interesting robustness property of batch normalization and the Adam optimizer, as well as a high sensitivity to the network structure.

Overall, the goal of this work is to draw attention to an overlooked factor in machine learning and neural networks:  nonsmoothness. The $\ReLU$ activation is probably the most widely used nonlinearity in this context, yet its nondifferentiability is mostly ignored. We highlight the fact that the default choice $\ReLU'(0) = 0$ seems to be the most robust, while different choices could potentially lead to instabilities. For a general nonsmooth nonlinearity, it is not clear \textit{a priori} which choice would be the most robust, if any, and our investigation underlines the potential importance of this question. Our research opens new directions regarding the impact of numerical precision on neural network training, its interplay with nonsmoothness and its combined effect with other learning factors, such as network architecture, batch normalization or optimizers. The main idea is that mathematically negligible factors are not necessarily computationally negligible.

\begin{ack}
The authors thank anonymous referees for constructive suggestions which greatly improved the paper. The authors acknowledge the support of the DEEL project, the AI Interdisciplinary Institute ANITI funding, through the French “Investing for the Future – PIA3” program under the Grant agreement ANR-19-PI3A-0004, Air Force Office of Scientific Research, Air Force Material Command, USAF, under grant numbers FA9550-19-1-7026, FA9550-18-1-0226, and ANR MaSDOL 19-CE23-0017-01.  J. Bolte  also acknowledges the support of ANR Chess, grant ANR-17-EURE-0010 and TSE-P.
\end{ack}
\bibliographystyle{abbrv}

\newpage

\appendix
\section{Mathematical details for Section~\ref{sec:mathAutoDiff}}

In Section~\ref{sec:additionalMath}, we  provide some elements of proof for Theorems~\ref{th:mainBP} and~\ref{th:noInfluenceOnSequence}. In Section \ref{sec:gijFullyConnected}, we explain how to check the assumptions of Definition \ref{def:reluNetwork} by describing the special case of fully connected ReLU networks.

\subsection{Elements of proof of Theorems \ref{th:mainBP} and \ref{th:noInfluenceOnSequence}}
\label{sec:additionalMath}
The proof arguments were described in \cite{bolte2020conservative,bolte2020mathematical}. We simply concentrate on justifying how the results described in these works apply to Definition \ref{def:reluNetwork} and point the relevant results leading to Theorems \ref{th:mainBP} and \ref{th:noInfluenceOnSequence}.

It can be inferred from Definition \ref{def:reluNetwork} that all elements in the definition of a $\ReLU$ network training problem are piecewise smooth, where each piece is an elementary $\logexp$ function. We refer the reader to \cite{scholtes2012introduction} for an introduction to piecewise smoothness and recent use of such notions in the context of algorithmic differentiation in \cite{bolte2020mathematical}. Let us first argue that the results of \cite{bolte2020mathematical} apply to Definition \ref{def:reluNetwork}.
\begin{itemize}
    \item We start with an explicit selection representation of $\backprop_s\  \ReLU$. Fix any $s \in \RR$ and consider the three functions $f_1 \colon x \mapsto 0$, $f_2 \colon x \mapsto x$ and $f_3 \mapsto s x$ with the selection index $t(x)= 1$ if $x<0$, $2$ if $x>0$ and $3$ if $x = 0$. We have for all $x$
    \begin{align*}
        f_{t(x)} = \ReLU(x)
    \end{align*}
    Furthermore, differentiating the active selection as in \cite[Definition 4]{bolte2020mathematical} we have
    \begin{align*}
        \hat{\nabla}^t f = 
        \begin{cases}
            0 \qquad \text{ if } x<0\\
            1 \qquad \text{ if } x>0\\
            s \qquad \text{ if } x=0
        \end{cases}
    \end{align*}
    and the right hand side is precisely the definition of $\backprop_s\  \ReLU$. This shows that we have a selection derivative as used in \cite{bolte2020mathematical}.
    \item Given a $\ReLU$ network training problem as in Definition \ref{def:reluNetwork}, we have the following property. 
    \begin{itemize}
        \item All elements in the $\ReLU$ network training problem are piecewise elementary $\logexp$. That is each piece can be identified with an elementary $\logexp$ function. Furthermore the selection process describing the choice of active function can similarly be described by elementary $\logexp$ functions with equalities and inequalities.
    \end{itemize}
\end{itemize}
Therefore, we meet the definition of $\logexp$ selection function in \cite{bolte2020mathematical} and all corresponding results apply to any $\ReLU$ network training problem as given in Definition \ref{def:reluNetwork}. Fix  $T \geq 1$, getting back to problem \eqref{eq:NNtraining}, using \cite[Definition 5]{bolte2020mathematical} and the selection derivative described above, for each $i=1,\ldots,N$, there is a conservative field $D_i \colon \RR^P \rightrightarrows \RR^P$ (definition of conservativity is given in \cite{bolte2020conservative} and largely described in \cite{bolte2020mathematical}) such that for any $s \in [0, T]$, and $\theta \in \RR^P$
\begin{align*}
    \backprop_s l_i(\theta) \in D_i(\theta).
\end{align*}
Using \cite[Corollary 5]{bolte2020conservative} we have $D_i(\theta) = \{\nabla l_i (\theta)\}$ for all $\theta$ outside of a finite union of differentiable manifolds of dimension at most $P-1$. This leads to Theorem \ref{th:mainBP} for $s \in [0,T]$. Theorem \ref{th:noInfluenceOnSequence} is deduced from the proof of \cite[Theorem 7]{bolte2020mathematical} (last paragraph of the proof) that with probability $1$, for all $k \in \NN$, for all $n = 1, \ldots, N$ and $s \in [0,T]$
\begin{align*}
    \backprop_s[\ell(f(x_n, \theta_{k,s}), y_n)] = \nabla_\theta \ell(f(x_n, \theta_{k,s}), y_n)
\end{align*}
since we have $\theta_{0,s} = \theta_0$ for all $s$, the generated sequence in \eqref{eq:SGDbackprop} does not depend on $s \in [0,T]$. This is Theorem \ref{th:noInfluenceOnSequence} for $s \in [0,T]$, note that a similar probabilistic argument was developped in \cite{bianchi2020convergence}. We may repeat the same arguments fixing $T<0$, so that both results actually hold for all $s \in [-T,T]$.

\subsection{The special case of fully connected  ReLU networks}
\label{sec:gijFullyConnected}
The functions $g_{i,j}$ in the composition~\eqref{eq:composition} can be described explicitly for any given neural network architecture. For the sake of clarity, we detail below the well-known case of fully connected  ReLU networks for multiclass classification. We denote by $K \geq 2$ the total number of classes.

Consider any fully connected  ReLU network architecture of depth $H$,  with the softmax function applied on the last layer. We denote by $d_h$ the size of each layer $h=1,\ldots,H$, and by $d_0$ the input dimension. In particular $d_H = K$ equals the number of classes. All the functions $f_{\theta}:\RR^{d_0} \to \RR^{d_H}$ represented by the network when varying the weight parameters $\theta \in \RR^P$ are of the form:
\[
f_{\theta}(x) = f(x,\theta) = \softmax \circ A_H \circ \sigma \circ A_{H-1} \circ \cdots \sigma \circ A_1 (x) \;,
\]
where each mapping $A_h:\RR^{d_{h-1}} \to \RR^{d_h}$ is affine (i.e., of the form $A_h(z)=W_h z + b_h$), where $\sigma(u) = \bigl(\ReLU(u_i)\bigr)_i$ applies the ReLU function component-wise to any vector~$u$, and where $\softmax(z) = \bigl(e^{z_i}/\sum_{k=1}^{d_H} e^{z_k}\bigr)_{1 \leq i \leq d_H}$ for any $z \in \RR^{d_H}$. The weight parameters $\theta \in \RR^P$ correspond to stacking all weight matrices $W_h$ and biases $b_h$ in a single vector (in particular, we have here $P=\sum_{h=1}^H d_h (d_{h-1}+1)$). In the sequel, we set $P_h = \sum_{j=h}^H d_j (d_{j-1}+1)$ and write $\theta_{h:H} \in \RR^{P_h}$ for the vector of all parameters involved from layer $h$ to the last layer $H$. We also write $\concatenate(x_1,\ldots,x_r)$ to denote the vector obtained by concatenating any $r$ vectors $x_1,\ldots,x_r$. In particular, we have $\theta_{h:H} = \concatenate(W_h,b_h,\theta_{h+1:H})$.

Note that the decomposition above took $x$ as input, not $\theta$. We now explain how to construct the $g_{i,j}$ in \eqref{eq:composition}. For each $i=1,\ldots,N$, the function $\theta \in \RR^P \mapsto f(x_i,\theta)$ can be decomposed as
\begin{equation}
f(x_i,\theta) = \softmax \circ g_{i,2H-1} \circ \ldots \circ g_{i,2} \circ g_{i,1} (\theta) \;,
\label{eq:networkdecomposed}
\end{equation}
where, roughly speaking, the $g_{i,2h-1}$ apply the affine mapping $A_h$ to the output $z_{h-1} \in \RR^{d_{h-1}}$ of layer $h-1$ and pass forward all parameters $\theta_{h+1:H} \in \RR^{P_{h+1}}$ to be used in the next layers, while the $g_{i,2h}$ apply the $\ReLU$ function to the first $d_h$ coordinates. More formally, $g_{i,1}:\RR^P \to \RR^{d_1 + P_2}$ is given by
\[
g_{i,1}(\theta) = \concatenate(W_1 x_i + b_1, \theta_{2:H}) \;,
\]
$g_{i,2}:\RR^{d_1 + P_2} \to \RR^{d_1 + P_2}$ maps any $(z_1,\theta_{2:H}) \in \RR^{d_1} \times \RR^{P_2}$ to
\[
g_{i,2}(z_1,\theta_{2:H}) = \concatenate\bigl(\sigma(z_1),\theta_{2:H}\bigr)
\]
and, for each layer $h = 2,\ldots,H$, the functions $g_{i,2h-1}:\RR^{d_{h-1}+P_h} \to \RR^{d_h+P_{h+1}}$ and $g_{i,2h}: \RR^{d_h+P_{h+1}} \to \RR^{d_h+P_{h+1}}$ are given by
\[
g_{i,2h-1}(z_{h-1},\theta_{h:H}) = \concatenate(W_h z_{h-1} + b_h,\theta_{h+1:H})
\]
and
\[
g_{i,2h}(z_h,\theta_{h+1:H}) = \concatenate\bigl(\sigma(z_h),\theta_{h+1:H}\bigr) \qquad \textrm{(for $h < H$).}
\]

Consider now the cross-entropy loss function $\ell:\Delta(K) \times \{1,\ldots,K\} \to \RR_+$ which compares any probability vector $q \in \Delta(K)$ of size $K$ (with non-zero coordinates $q_i >0$) with any true label $y \in \{1,\ldots,K\}$, given by
\[
\ell(q,y) = -\log q(y) \;.
\]

Finally, using \eqref{eq:networkdecomposed}, the functions $l_i \colon \RR^P \to \RR$ appearing in \eqref{eq:NNtraining}-\eqref{eq:composition} can be decomposed as
\[
l_i(\theta) = \ell\bigl(f(x_i, \theta), y_i\bigr) = \bigl(q \in \Delta(K) \mapsto \ell(q,y_i)\bigr) \circ \softmax \circ g_{i,2H-1} \circ \ldots \circ g_{i,2} \circ g_{i,1} (\theta) \;.
\]

The last decomposition satisfies \eqref{eq:composition} with $M=2H+1$. Since $\ell$ is the cross-entropy loss function, all $M$ functions involved in this decomposition are either elementary {\logexp} or consist in applying $\ReLU$ to some coordinates of their input, and they are all locally Lipschitz, as required in Definition~\ref{def:reluNetwork}. This provides an explicit description of fully connected ReLU network and a similar description can be done for all architectures studied in this work.

\section{First experiment in 64 bits precision, and using a different activation}
\label{sec:firstExp64BitsPrec}
The code and results associated with all experiments presented in this work are publicly available here: \url{https://github.com/deel-ai/relu-prime}.

\begin{figure}[h]
\centering
    
      \begin{center}
        \includegraphics[width=.31\textwidth]{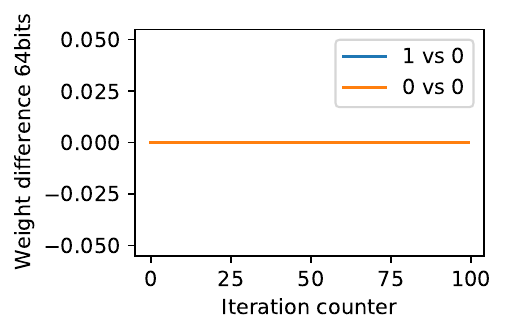} \quad \includegraphics[width=.31\textwidth]{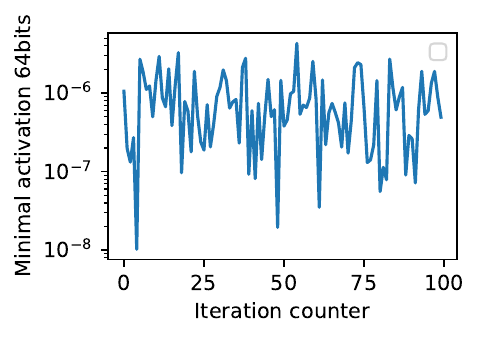} \quad
        \includegraphics[width=.3\textwidth]{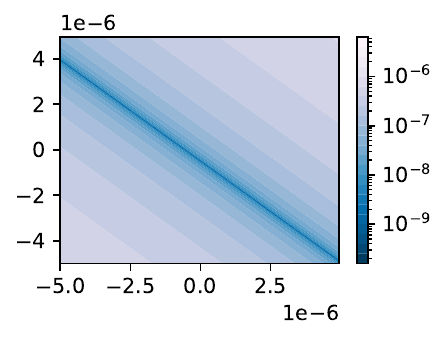} 
      \end{center}
    \caption{Same experiment as Figure~\ref{fig:first_xp} in 64 bits precision. Left: Difference between network parameters ($L^1$ norm), 100 iterations within an epoch. ``0 vs 0''  indicates $\|\theta_{k,0} - \tilde{\theta}_{k,0}\|_1$ where $\tilde{\theta}_{k,0}$ is a second run for sanity check, ``0 vs 1'' indicates $\|\theta_{k,0} - \theta_{k,1}\|_1$. Center: minimal absolute activation of the hidden layers within the $k$-th mini-batch, before $\ReLU$. At iteration 65, there is no jump on the left and no drop in the center anymore. Right: illustration of the bifurcation zone at iteration $k=65$ (same weight parameter plane as in Figure~\ref{fig:first_xp}, but in 64 bits precision). The quantity represented is the absolute value of the neuron of the first hidden layer which was exactly zero in 32 bits (see Figure~\ref{fig:first_xp}) before application of $\ReLU$. Exact zeros are represented in white.}
    \label{fig:first_xp_64bits}
\end{figure}

\paragraph{64 bits precision.} We reproduce the same bifurcation experiment as in Section \ref{sec:prelimMNIST} under 64 bits  arithmetic precision. The results are represented in Figure \ref{fig:first_xp_64bits} which is to be compared with its 32 bits counterpart in Figure~\ref{fig:first_xp}. As mentioned in the main text, the bifurcation does not occur anymore. Indeed the magnitude of the smallest activation before application of $\ReLU$ is of the same order, but this time it is well above machine precision which is around $10^{-16}$. When depicting the same neighborhood as in Figure \ref{fig:first_xp}, the effect of numerical error completely disappears, the bifurcation zone being reduced to a segment in the picture, which is consistent with Theorems \ref{th:mainBP} and \ref{th:noInfluenceOnSequence}. 

\begin{figure}[h]
\centering
    
      \begin{center}
        \includegraphics[width=.31\textwidth]{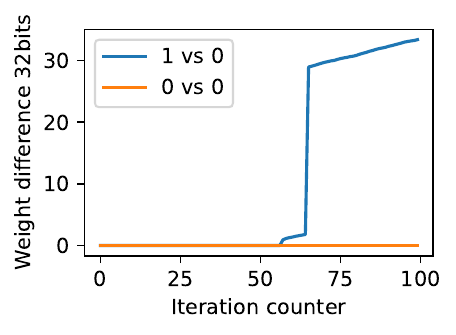} \quad \includegraphics[width=.31\textwidth]{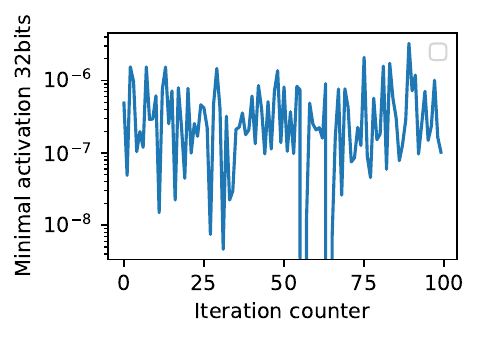} \\
        \includegraphics[width=.3\textwidth]{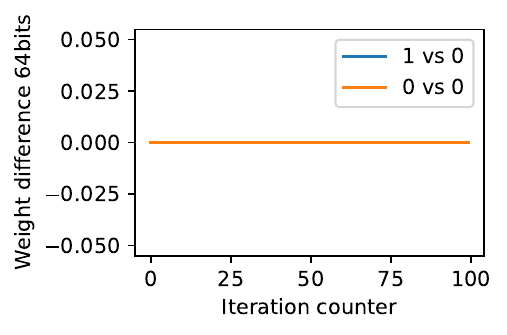} \quad
        \includegraphics[width=.3\textwidth]{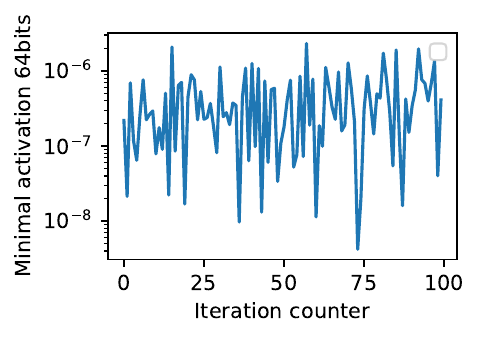} 
      \end{center}
    \caption{Same experiment as Figure~\ref{fig:first_xp} with $\ReLU6$ in place of $\ReLU$. Top: 32 bits weight difference and minimal activation before application of $\ReLU6$. Bottom: 64 bits weight difference and minimal activation before application of $\ReLU6$}
    \label{fig:first_xp_relu6}
\end{figure}

\paragraph{ReLU6 activation.} We conducted the same experiment with the $\ReLU6$ activation function in place of $\ReLU$ and found similar results on a slightly larger network ($754$, $4000$, $256$). Recall that $\ReLU6$ is equal to $\ReLU$ for $x< 6$ and equal to $6$ for $x\geq 6$ and the default choice of derivatives at non differentiable points are zero. The illustration is given in Figure \ref{fig:first_xp_relu6}.

\section{Details on Monte Carlo sampling in Table \ref{table:precision}}
\label{sec:montecarlo}
The code and results associated with all experiments presented in this work are publicly available here: \url{https://github.com/deel-ai/relu-prime}.

Recall that we want to estimate the relative volume of the set
\begin{align*}
    S_{01} = \{\theta \in \RR^P :\ \exists i \in \{1,\ldots, N\},\  \backprop_0[l_i](\theta) \neq \backprop_1[l_i](\theta)\}\subset S.
\end{align*}
by Monte Carlo sampling. We randomly generate a set of parameters $\{\theta_j\}_{j=1}^M$, with $M=1000$, for a fully connected network architecture $f$ composed of $L$ hidden layers using Kaiming-Uniform \cite{he2015delving} random weight generator. Given this sample of parameters, iterating on the whole MNIST dataset, we approximate the proportion of $\theta_j$ for which $ \backprop_0(l_i)(\theta_j) \neq \backprop_1(l_i)(\theta_j)$ for some $i$, for different networks and under different conditions. More precisely, denoting by $Q$ the number of mini-batches considered in the MNIST dataset, and by $B_q \subset \{1,\ldots N\}$ the indices corresponding to the mini-batch $q$, for $q = 1,\ldots Q$,  the first line of Table \ref{table:precision} is given by the formula
\begin{align*}
    \frac{1}{M} \sum_{m=1}^M \mathbb{I}\left( \exists q \in \{1,\ldots, Q\},\, \backprop_0\left[\sum_{j \in B_q} \l_j(\theta_m) \right]\neq \backprop_1\left[\sum_{j \in B_q} \l_j(\theta_m)\right]\right),
\end{align*}
where the function $\mathbb{I}$ takes value $1$ or $0$ depending on the validity of the statement in its argument. Similarly, the second line of Table \ref{table:precision} is given by the formula
\begin{align*}
    \frac{1}{MQ} \sum_{m=1}^M\sum_{q = 1}^Q \mathbb{I}\left( \backprop_0\left[\sum_{j \in B_q} \l_j(\theta_m) \right]\neq \backprop_1\left[\sum_{j \in B_q} \l_j(\theta_m)\right]\right),
\end{align*}
while the last line provides statistics of the quantity
\begin{align*}
    \frac{\left\|\backprop_0\left[\sum_{j \in B_q} \l_j(\theta_m) \right] - \backprop_1\left[\sum_{j \in B_q} \l_j(\theta_m)\right]\right\|}{\left\|\backprop_0\left[\sum_{j \in B_q} \l_j(\theta_m) \right]\right\|},
\end{align*}
conditioned on $q, m$ being such that $\backprop_0\left[\sum_{j \in B_q} \l_j(\theta_m) \right]\neq \backprop_1\left[\sum_{j \in B_q} \l_j(\theta_m)\right]$.

The error margin associated with the confidence interval on the first line of Table \ref{table:precision} is computed using Hoeffding's inequality at risk level $5\%$. It is given by the formula 
\begin{align*}
    \sqrt{\frac{ \ln\left( \frac{2}{0.05}\right)}{2M}} \;.
\end{align*}
As for the confidence interval of the second line of Table~\ref{table:precision}, we use the bounded differences inequality (a.k.a. McDiarmid's inequality) at risk level $5\%$. The error margin is given by the formula
\begin{align*}
    \sqrt{\frac{1}{2} \left( \frac{1}{M} + \frac{1}{Q} \right) \ln\left( \frac{2}{0.05}\right)} \;.
\end{align*}

\section{Complements on experiments}
\label{sec:additionalExpe}
The code and results associated with all experiments presented in this work are publicly available here: \url{https://github.com/deel-ai/relu-prime}.

\subsection{Benchmark datasets and architectures}
\label{sec:benchmarksAndArchitectures}
Overview of the datasets used in this work. These are image classification benchmarks, the corresponding references are respectively \cite{lecun1998gradient,krizhevsky2010cifar,netzer2011reading}.
\begin{center}
    \begin{tabular}{cccc}
    Dataset & Dimensionality & Training set & Test set \\\hline
    MNIST& $28 \times 28$ (grayscale)  & 60K & 10K \\
    CIFAR10& $32 \times 32$ (color) &   60K& 10K\\
    SVHN&$32 \times 32$ (color) &   600K& 26K\\
    ImageNet & $224 \times 224$ (color) &   1300K & 50K\\
    \end{tabular}
\end{center}

Overview of the neural network architectures used in this work. The corresponding references are respectively \cite{srivastava2014dropout,simonyan15very,he2016deep}.
\begin{center}
    \begin{tabular}{cccc}
    Name & Type & Layers & Loss function \\\hline
    Fully connected& fully connected  & 4 & Cross-entropy \\
    VGG11& convolutional &9& Cross-entropy\\
    ResNet18&convolutional &18& Cross-entropy\\
    ResNet50&convolutional &50& Cross-entropy\\
    \end{tabular}
\end{center}

\paragraph{Fully connected architecture:} This architecture corresponds to the one used in \cite{srivastava2014dropout}. We only trained this network on MNIST, the resulting architecture has an input layer of size 784, three hidden layers of size 2048 and the ouput layer is of size 10.

\paragraph{VGG11 architecture:} We used the implementation proposed in the following repository \url{https://github.com/kuangliu/pytorch-cifar.git} which adapts the VGG11 implementation of the module \texttt{torchvision.models} for training on CIFAR10. The only modification compared to the standard implementation is the fully connected last layers which only consist in a linear $512 \times 10$ layer. When adding batch normalization layers, it takes place after each convolutional layer. 

\paragraph{ResNet18 architecture:} We use PyTorch implementation for this architecture found in the module \texttt{torchvision.models}. We only modified the size of the output layer (10 vs 1000), the size of the kernel in the first convolutional layer (3 vs 7) and replaced batch normalization layers by the identity (when we did not use batch normalization).

\subsection{Additional Experiments with MNIST and fully connected networks}
\label{sec:additionalMNIST}
\begin{figure}[h]
\centering
      \begin{center}
        \includegraphics[width=\textwidth]{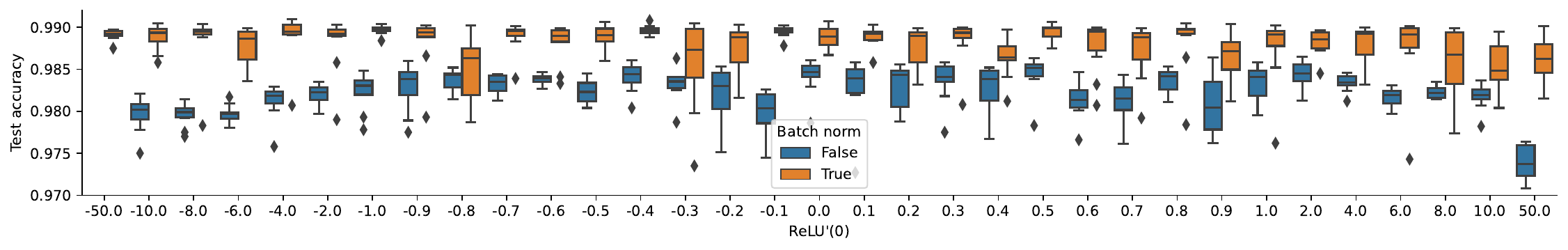} 
        \includegraphics[width=.38\textwidth]{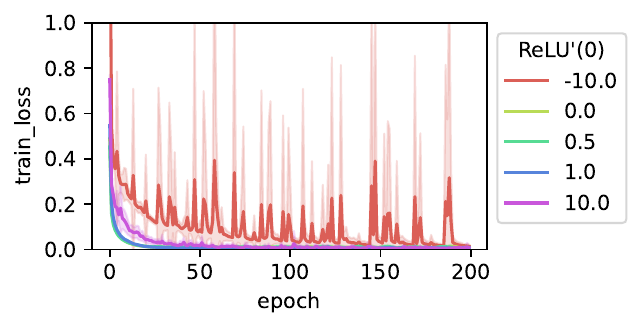} \quad\includegraphics[width=.28\textwidth]{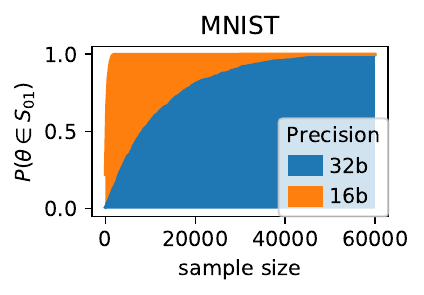}\quad\includegraphics[width=.28\textwidth]{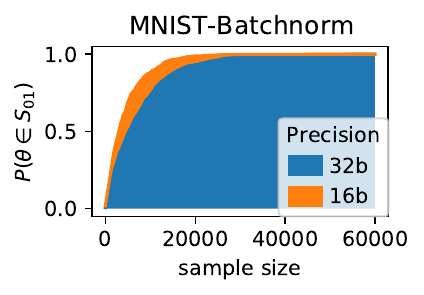}
      \end{center}
    \caption{Top: Test error on MNIST with a fully connected  784-2048-2048-2048-10 network. The boxplots and shaded areas represent variation over ten random initializations. We recover the bell shaped curve, but the sensitivity to $\ReLU'(0)$ is less important. Bottom left: corresponding training loss, higher magnitude of $\ReLU'(0)$ induces chaotic oscillation explaining the decrease in test accuracy. Bottom center and right: relative volume estimation of the bifurcation zone without and with batch normalization. Batch normalization increases the size of the bifurcation zone with 32 bits arithmetic and decreases it under 16 bits arithmetic precision.}
    \label{fig:MNISTTest}
\end{figure}

\begin{figure}[h]
\centering
      \begin{center}
        \includegraphics[width=\textwidth]{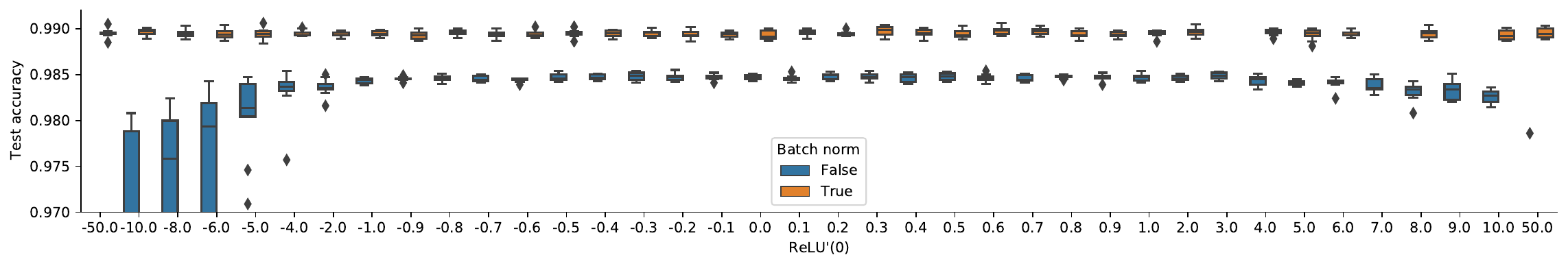} 
      \end{center}
    \caption{Same experiment as in Figure \ref{fig:MNISTTest} without learning rate tuning.}
    \label{fig:MNISTTestNoOptuna}
\end{figure}

We conducted the same experiments as in Section \ref{sec:numMainEffect} with a fully connected  784-2048-2048-2048-10 network on MNIST. The results are represented in Figure \ref{fig:MNISTTest} which parallels the results in Figure~\ref{fig:cifarVGG} on VGG11 with CIFAR10. We observe a similar qualitative behavior, but the fully connected architecture is less sensitive to the magnitude chosen for $\ReLU'(0)$. Note that in this case, learning rate tuning with optuna \cite{akiba2019optuna} induces a lot of spurious variability. Indeed, the same experiment with fixed learning rate results in a much smoother bell shape in Figure \ref{fig:MNISTTestNoOptuna}.

\begin{figure}[h]
\centering
      \begin{center}
        \includegraphics[width=.5\textwidth]{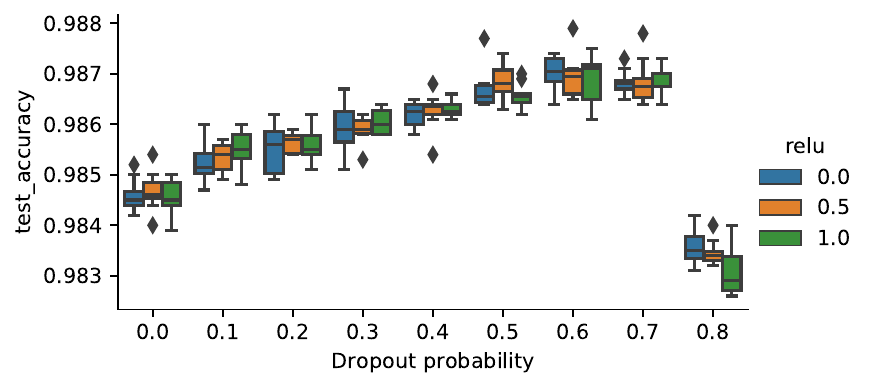} 
      \end{center}
    \caption{Experiment on combination of the choice of $\ReLU'(0)$ with dropout on MNIST with a fully connected 784-2048-2048-2048-10 network. The boxplots represent 10 random initializations.}
    \label{fig:MNISTDropout}
\end{figure}

We investigated further the effect of combining different choices of $\ReLU'(0)$ with dropout \cite{srivastava2014dropout}. Dropout is another algorithmic way to regularize deep networks and it was natural to wonder if it could have a similar effect as batch normalization. Using the same network, we combined different choices of dropout probability with different choices of $\ReLU'(0)$. The results are represented in Figure \ref{fig:MNISTDropout} and suggests that dropout has no conjoint effect.

\subsection{Additional experiments with VGG11}
\label{sec:additionalVGG11}

This section complements Sections~\ref{sec:numMainEffect} and~\ref{sec:numBatchNorm}, with additional experiments with VGG11.

\begin{figure}[h]
\centering
  \begin{center}
    \includegraphics[width=.45
    \textwidth]{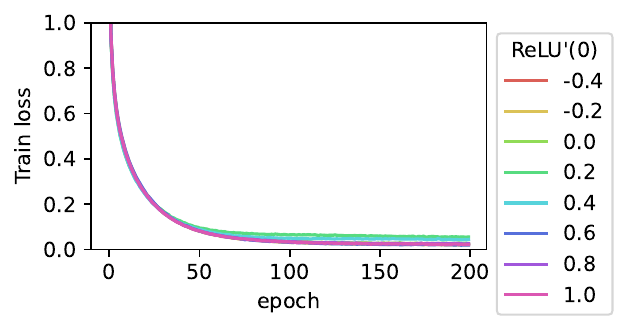} \qquad
    \includegraphics[width=.45\textwidth]{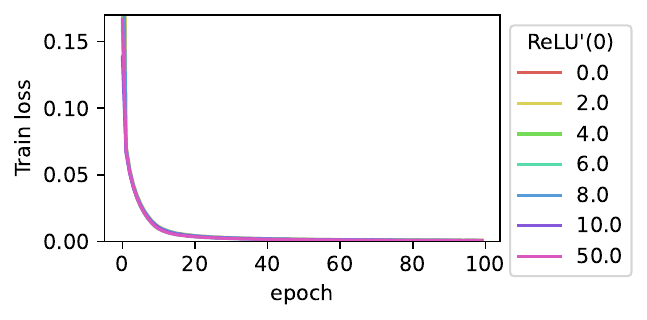}
  \end{center}
  \caption{Training loss on CIFAR10 with VGG11 (left) and SVHN with VGG11 (right). The instability induced by the choice of $\ReLU'(0)$ completely disappears with batch normalization.}
\label{fig:trainWithBatchNorm}
\end{figure}

\paragraph{Batch normalization.} As suggested by the experiment shown in Section \ref{sec:numBatchNorm}, batch normalization stabilizes the choice of $\ReLU'(0)$, leading to higher test performances. We display in Figure \ref{fig:trainWithBatchNorm} the decrease of training loss on CIFAR 10 and SVHN, for VGG11 with batch normalization. We see that the choice of $\ReLU'(0)$ has no impact and that the chaotic oscillations induced by this choice have completely disappeared.

\begin{figure}[h]
\centering
  \begin{center}
    \includegraphics[width=.53\textwidth]{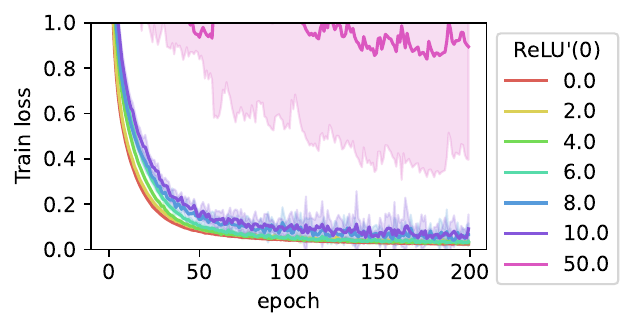}
    \quad
    \includegraphics[width=.43\textwidth]{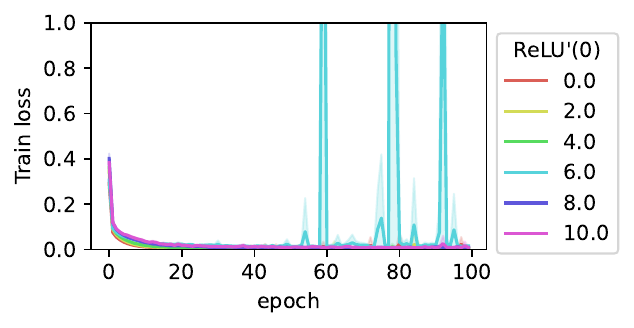}
  \end{center}
  \caption{Training losses on CIFAR10 (left) and SVHN (right) on VGG network trained with Adam  optimizer. The filled area represent standard deviation over ten random initializations. }
\label{fig:cifarVGGAdam}
\end{figure}

\paragraph{Adam optimizer.} The training curves corresponding to Figure \ref{fig:SVHNVGG} are shown in Figure \ref{fig:cifarVGGAdam}. They suggest that the Adam  optimizer features much less sensitivity than SGD to the choice of $\ReLU'(0)$. This is seen with a relatively efficient buffering effect on the induced oscillatory behavior on training loss decrease.

\begin{figure}[h]
\centering
  \begin{center}
    \includegraphics[width=.35\textwidth]{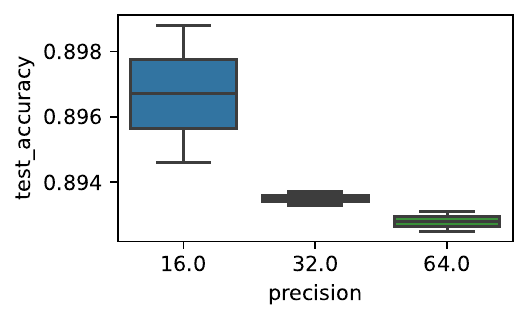}
    \qquad
    \includegraphics[width=.35\textwidth]{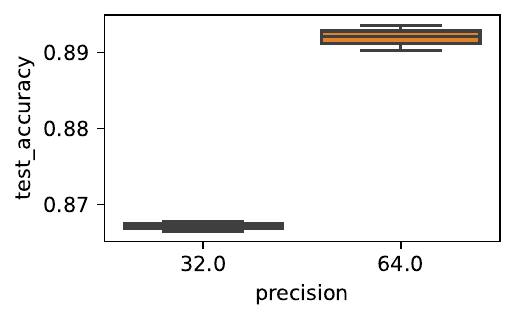}
  \end{center}
  \caption{Test accuracy for different numerical precisions with a VGG11 network on CIFAR10. Left: $\ReLU'(0)=0$. Right: $\ReLU'(0)=1$.}
\label{fig:cifarVGGPrecision}
\end{figure}

\paragraph{Numerical precision.} For this neural network we investigated the joint effect of $\ReLU'(0)$ and numerical precision (16, 32 or 64 bits). The results are displayed in Figure \ref{fig:cifarVGGPrecision}. The choice $\ReLU'(0) = 1$ leads to such a high instability in 16 bits precision that we were not able to tune the learning rate to train the network without explosion of the weights. In 32 bits, a few experiments resulted in non-convergent training---these were removed. We observe first that for $\ReLU'(0) = 0$ numerical precision has barely any effect while for $\ReLU'(0) = 1$ it leads to an increase in test accuracy. Furthermore, we observe that $\ReLU'(0) = 1$ with 64 bits precision leads to the same test accuracy as $\ReLU'(0) = 0$ in 32 bits precision.

\subsection{Additional experiments with ResNet18}
\label{sec:additionalResnet}
\begin{figure}[h]
\centering
  \begin{center}
    \includegraphics[width=.52
    \textwidth]{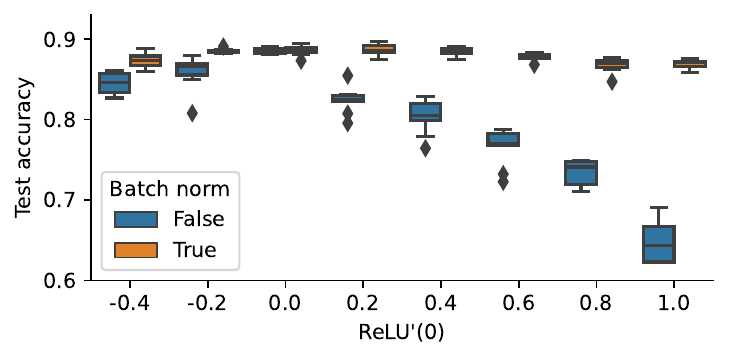} \qquad
    \includegraphics[width=.35
    \textwidth]{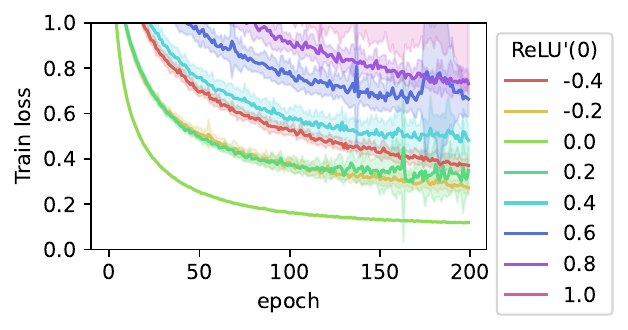} \quad
    \includegraphics[width=.55
    \textwidth]{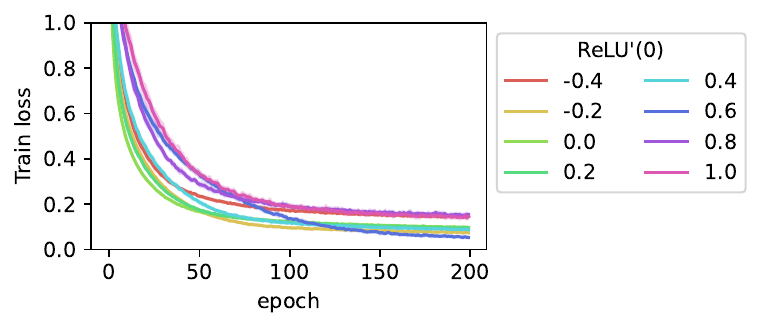} 
  \end{center}
  \caption{Training experiment on CIFAR10 with Resnet18 and the SGD optimizer. Top left: test accuracy with and without batch normalization. Top right: training loss during training without batch normalization. Bottom: training loss during training with batch normalization.}
\label{fig:cifarResNet}
\end{figure}

We performed the same experiments as the ones described in Section \ref{sec:consequencesLearning} using a ResNet18 architecture trained on CIFAR 10. The test error, training loss evolution with or without batch normalization are represented in Figure \ref{fig:cifarResNet}. We have similar qualitative observations as with VGG11. We note that the ResNet18 architecture is much more sensitive to the choice of $\ReLU'(0)$:
\begin{itemize}
    \item Test performances degrade very fast. Actually, beyond a magnitude of 0.2, we could not manage to train the network without using batch normalization.
    \item Even when using batch normalization, the choice of $\ReLU'(0)$ seems to have an effect for relatively small variations. This is qualitatively different from what we observed with VGG11 and fully connected architectures. 
\end{itemize}

Similar Monte Carlo relative volume experiments were carried out for this network architecture; the results are presented in Figure \ref{fig:cifarResNetVol}. The results are qualitatively similar to what we observed for the VGG11 architecture: the bifurcation zone is met very often for 16 bits precision, and the addition of batch normalization increases this frequency in 32 bits precision. Note that we did not observe a significant variation in 16 bits precision.

\begin{figure}[h]
\centering
  \begin{center}
    \includegraphics[width=.3
    \textwidth]{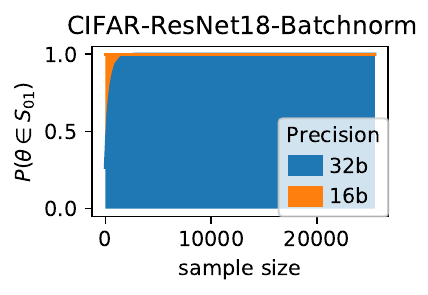} \qquad \includegraphics[width=.3
    \textwidth]{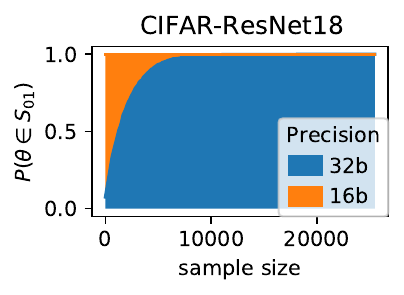}
  \end{center}
  \caption{Relative volume Monte Carlo estimation on CIFAR10 with Resnet18 with and without batch normalization under 16 bits or 32 bits precision.}
\label{fig:cifarResNetVol}
\end{figure}

\subsection{Additional experiments with ResNet50 on ImageNet}
\label{sec:additionalResnet}

\begin{figure}[h]
\centering
  \begin{center}
    \includegraphics[width=.3\textwidth]{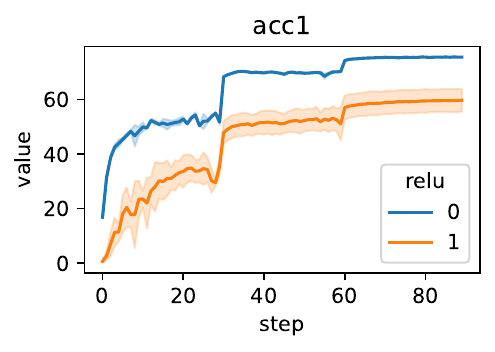} 
  \end{center}
  \caption{Test accuracy during training of a Resnet50 on ImageNet with \ed{SGD}. The shaded area represents two runs. We can see a massive drop in test accuracy with $\ReLU'(0) = 1$.}
\label{fig:cifarResNetVol}
\end{figure}

\section{Complimentary information, total amount of compute and resources used}
\label{app:repro}

All the experiments were run on a 2080ti GPU.
 The code corresponding to the experiments and experiments results are available at \url{https://github.com/deel-ai/relu-prime}
 Details about each test accuracy experiments are reported on Table \ref{table:experiments_details}. CIFAR10 is released under MIT license, MNIST, SVHN and R are released under GNU general public license, ImageNet is released under BSD license Numpy and pytorch are released under BSD license, python is released under the python sofware fondation license. 

\begin{table}
\begin{center}

 \begin{tabular}{||l | l | l| l | l| l | l | l ||} 
 \hline
 Dataset & Network & Optimizer & Batch size & Epochs & Time by epoch & Repetitions\\
 \hline
 \hline
 CIFAR10 & VGG11 & SGD & 128 & 200 & 9 seconds & 10 times  \\  
 \hline
 CIFAR10 & VGG11 & Adam & 128  & 200 & 10 seconds & 10 times \\  
 \hline
 CIFAR10 & ResNet18 & SGD & 128  & 200 & 13 seconds & 10 times \\  
 \hline
 SVHN & VGG11 & Adam & 128 & 64 & 85 seconds & 10 times  \\  
 \hline
 MNIST & MLP & SGD & 128 & 200 & 2 seconds & 10 times \\  
 \hline
 
\end{tabular}

\caption{Experimental setup}
 \label{table:experiments_details}
\end{center}
\end{table}
\end{document}